\documentclass[sigconf,nonacm]{acmart}

\usepackage{booktabs}
\usepackage{multirow}
\usepackage{makecell}
\usepackage{arydshln}
\usepackage{graphicx}
\usepackage{subcaption}
\usepackage{caption}
\usepackage{tabularx}
\usepackage{adjustbox}
\usepackage{threeparttable} 
\AtBeginDocument{%
  }

\setcopyright{none}
\renewcommand\footnotetextcopyrightpermission[1]{}

\begin{document}

\clearpage

\title{CTR-Sink: Attention Sink for Language Models in Click-Through Rate Prediction}

\author{Zixuan Li}
\orcid{0009-0002-5595-2343}
\authornote{This work was done during the internship at Ant Group.}
\affiliation{%
  \institution{NLPR, Institute of Automation, Chinese Academy of Sciences}
  \city{Beijing}
  \country{China}
}
\email{zixuan.li@nlpr.ia.ac.cn}

\author{Binzong Geng}
\orcid{0009-0003-4382-0543}
\affiliation{%
  \institution{Ant Group}
  \city{Hangzhou}
  \country{China}}
\email{gengbinzong.gbz@antgroup.com}

\author{Jing Xiong}
\orcid{0000-0003-2986-6978}
\affiliation{%
  \institution{The University of Hong Kong}
  \city{Hong Kong}
  \country{China}
}
\email{junexiong@connect.hku.hk}

\author{Yong He}
\orcid{0009-0000-5390-2655}
\affiliation{%
  \institution{Ant Group}
  \city{Hangzhou}
  \country{China}}
\email{heyong.h@antgroup.com}

\author{Yuxuan Hu}
\orcid{0009-0005-8571-118X}
\affiliation{%
  \institution{City University of Hong Kong}
  \city{Hong Kong}
  \country{China}
}
\email{yuxuanhu7-c@my.cityu.edu.hk}

\author{Jian Chen}
\orcid{0000-0002-4570-2271}
\affiliation{%
  \institution{The University of Hong Kong}
  \city{Hong Kong}
  \country{China}
}
\email{ccccccj03@connect.hku.hk}

\author{Dingwei Chen}
\orcid{0009-0007-0633-8896}
\affiliation{%
  \institution{Sun Yat-sen University}
  \city{Guangzhou}
  \country{China}
}
\email{chendw26@mail2.sysu.edu.cn}

\author{Xiyu Chang}
\orcid{0009-0005-9109-9193}
\affiliation{%
  \institution{Ant Group}
  \city{Hangzhou}
  \country{China}}
  \email{changxiyu.cxy@antgroup.com}

\author{Ngai Wong}
\orcid{0000-0002-3026-0108}
\affiliation{%
  \institution{The University of Hong Kong}
  \city{Hong Kong}
  \country{China}
}
\email{nwong@eee.hku.hk}

\author{Liang Zhang}
\orcid{0000-0002-7744-7789}
\affiliation{%
  \institution{Ant Group}
  \city{Hangzhou}
  \country{China}}
  \email{zhuyue.zl@antgroup.com}

\author{Linjian Mo}
\orcid{0000-0002-6682-1448}
\affiliation{%
  \institution{Ant Group}
  \city{Hangzhou}
  \country{China}}
\email{linyi01@antgroup.com}

\author{Chengming Li}
\orcid{0000-0002-4592-3875}
\authornote{The corresponding authors.}
\affiliation{%
  \institution{Shenzhen MSU-BIT University}
  \city{Shenzhen}
  \country{China}}
\email{licm@smbu.edu.cn}

\author{Chuan Yuan}
\authornotemark[2]
\affiliation{%
  \institution{Ant Group}
  \city{Hangzhou}
  \country{China}}
  \email{yuanzheng.xy@antgroup.com}

\author{Zhenan Sun}
\authornotemark[2]
\affiliation{%
  \institution{NLPR, Institute of Automation, Chinese Academy of Sciences}
  \city{Beijing}
  \country{China}
}
\email{znsun@nlpr.ia.ac.cn}

\renewcommand{\shortauthors}{Zixuan Li et al.}

\begin{abstract}

Click-Through Rate (CTR) prediction, a core task in recommendation systems, estimates user click likelihood using historical behavioral data. Modeling user behavior sequences as text to leverage Language Models (LMs) for this task has gained traction, owing to LMs’ strong semantic understanding and contextual modeling capabilities. However, a critical structural gap exists: user behavior sequences consist of discrete actions connected by semantically empty separators, differing fundamentally from the coherent natural language in LM pre-training. This mismatch causes \textit{semantic fragmentation}, where the lack of syntactic coherence causes LM attention to scatters across irrelevant tokens instead of focusing on meaningful behavior boundaries and inter-behavior relationships, degrading prediction performance.
To address this, we propose \textsc{CTR-Sink}, a novel framework introducing behavior-level attention sinks tailored for recommendation scenarios. Inspired by attention sink theory, it constructs attention focus sinks and dynamically regulates attention aggregation via external information. Specifically, we insert sink tokens between consecutive behaviors, incorporating recommendation-specific signals such as temporal distance to serve as stable attention sinks.
To enhance generality, we design a two-stage training strategy that explicitly guides LM attention toward sink tokens and an attention sink mechanism that amplifies inter-sink dependencies to better capture behavioral correlations. Experiments on one industrial dataset and two open-source datasets (MovieLens, Kuairec), alongside visualization results, validate the method's effectiveness across scenarios.
\end{abstract}




\maketitle
\begingroup\small\noindent\raggedright\textbf{Publication Note:}
Accepted by the 32nd ACM SIGKDD Conference on Knowledge Discovery and Data Mining (KDD 2026).
\par\endgroup

\newcommand\kddavailabilityurl{https://doi.org/10.5281/zenodo.20408470}
\ifdefempty{\kddavailabilityurl}{}{
\begingroup\small\noindent\raggedright\textbf{Resource Availability:}\\
The source code of this paper has been made publicly available at
\url{\kddavailabilityurl}. The GitHub repository is available at
\url{https://github.com/UGUESS-lzx/CTR-SINK}.
\endgroup
}
\setcounter{page}{1}

\section{Introduction}

Click-Through Rate (CTR), as a key metric for measuring user responsiveness to recommendations, advertisements, or search results, predicts the likelihood of users clicking on items based on their historical behavioral data. It plays a crucial role in recommendation systems and online advertising. 

\begin{figure}[h!]
  \centering

  \subcaptionbox{Attention visualization on coherent natural language data.\label{fig:subfig2}}{%
    \includegraphics[width=0.2\textwidth]{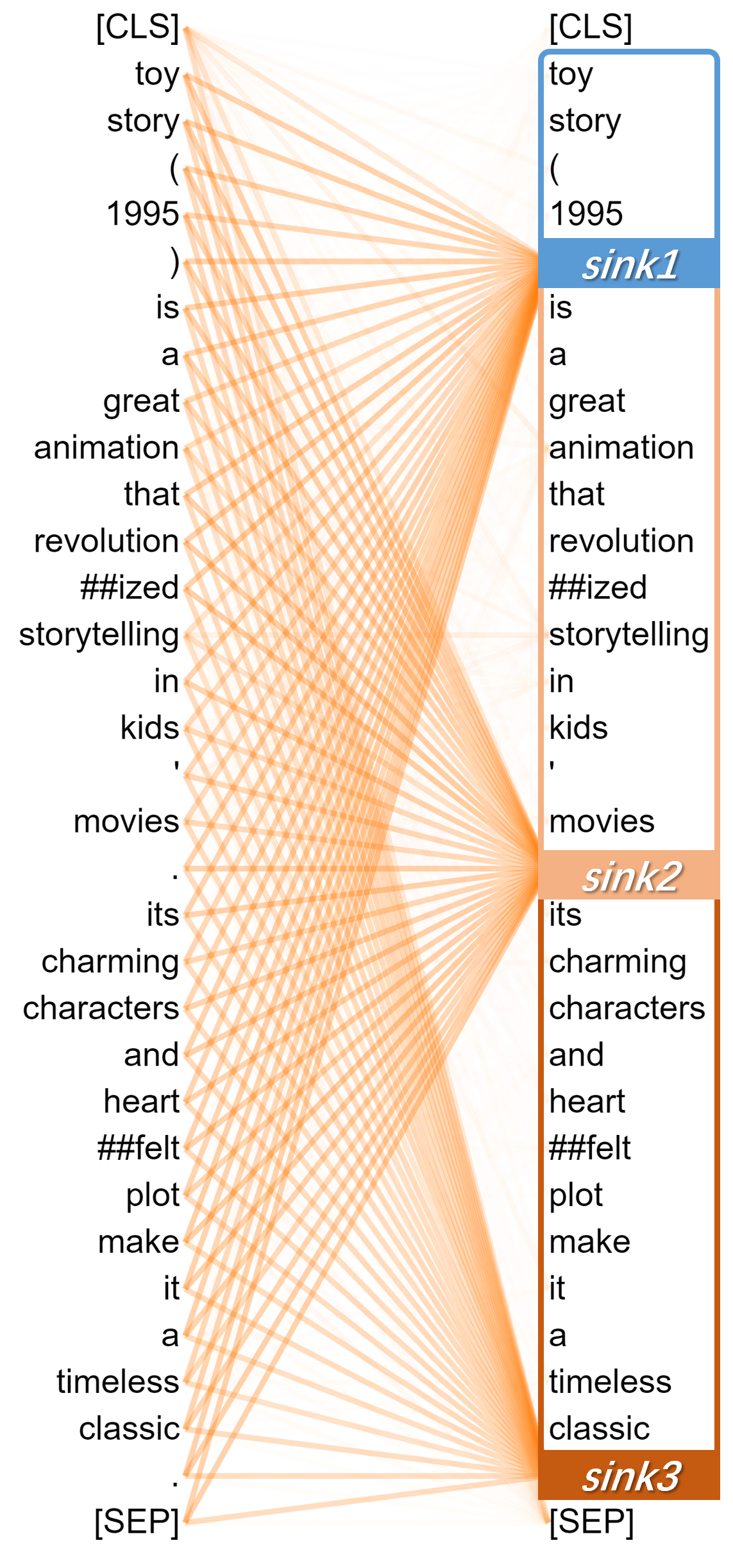}
  }
  \hfill
  \subcaptionbox{Attention visualization on textual user behaviors.\label{fig:subfig3}}{%
    \includegraphics[width=0.2\textwidth]{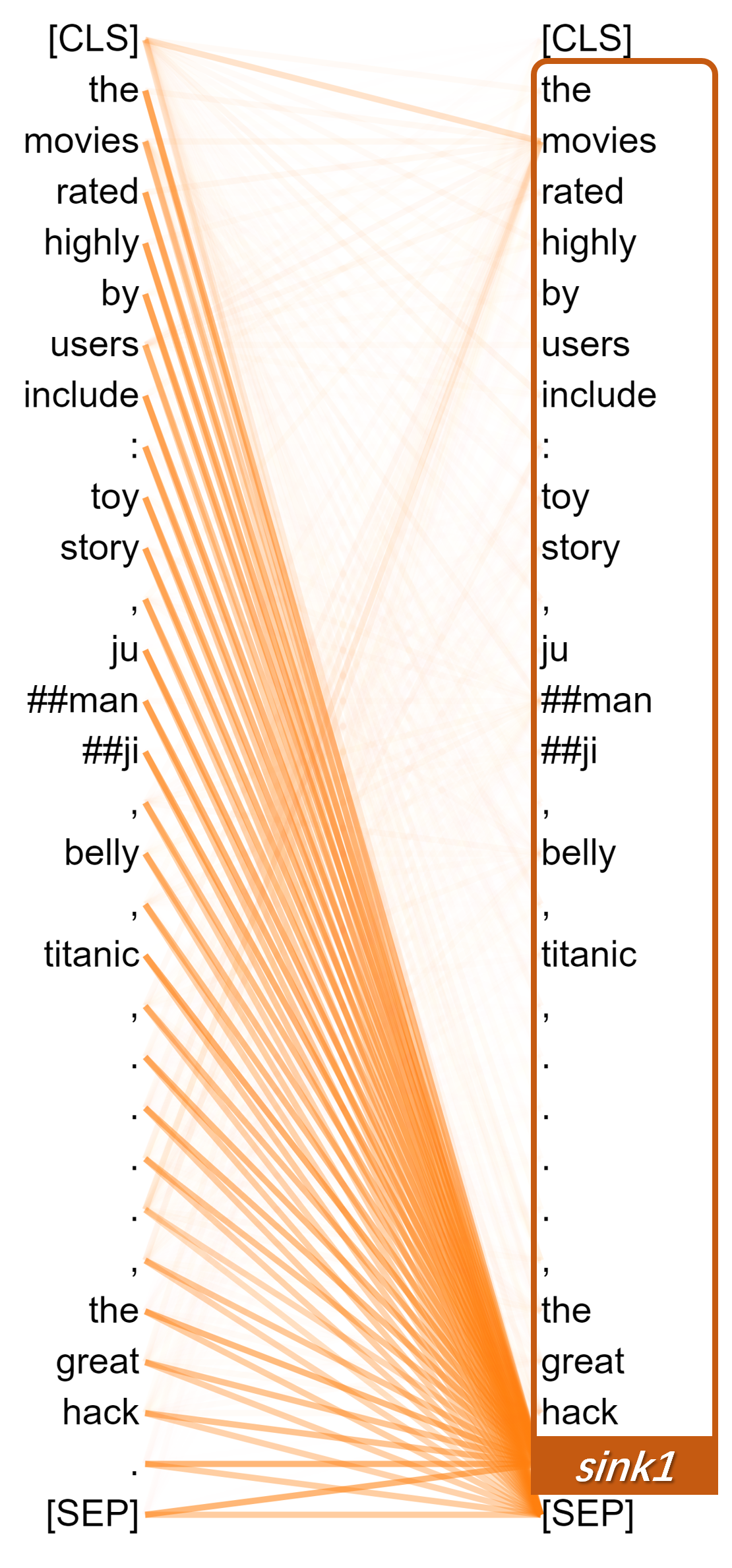}
  }
  
  \caption{Visualization of Semantic Fragmentation. It can be seen that the coherent natural language in (a) concentrates attention through natural convergence points, while in (b), textual user behaviors suffer from scattered attention due to semantic fragmentation.}
  \label{fig:combined}
\end{figure}

Inspired by the remarkable success of language models(LMs) across multiple domains in recent years\cite{devlin2019bert,zhao2023survey,radford2019language,mann2020language}, numerous studies have begun to explore the integration of LMs into recommendation tasks. Early works adopt In-Context Learning\cite{mann2020language}, using natural language prompts to ask LMs for recommendations directly\cite{zhang2021language,dai2023uncovering,gao2023chat}. Recently, some works focus on using LMs for semantic modeling of user behaviors, such as M6-Rec\cite{cui2022m6}, Tallrec\cite{bao2023tallrec}, BAHE\cite{geng_breaking_2024}, CTRL\cite{li_ctrl_2023}, CoLLM\cite{zhang_collm_2025}, TBHI\cite{chen_tbin_2023}, MSD\cite{zhang_balancing_2025}.
A common practice in these studies involves directly converting information such as users' historical behavior sequences into textual descriptions, which are then fed into LMs to extract semantic features for downstream tasks.

However, these methods implicitly assume that LMs can automatically capture the semantics of behavior sequences, overlooking a critical discrepancy: user behavior sequences in recommendation scenarios are fundamentally different from the natural language data used for LM pre-training. LM pre-training data (e.g., movie reviews) consists of coherent text with inherent logical structures (e.g., subject-verb-object relationships, causal connections), enabling meaningful semantic expression. In contrast, user behavior sequences are concatenations of discrete actions(e.g., Movie A, Movie B, Movie C, ...), connected only by semantically empty separators (e.g., commas) and lacking grammatical structure\cite{lin_rella_2024,geng_breaking_2024,wang_flip_2024,chang_singleton_2025}. As visualized in Figure~\ref{fig:combined}, this structural gap leads to \textit{semantic fragmentation}—LMs' attention  is concentrated on the last token while being notably scattered across all other tokens, failing to focus on key tokens that define behavior boundaries or inter-behavior relationships. And such attention aggregation has been proven to help language models better perform context modeling\cite{luo_bge_2024,xiaoefficient,wang_label_2023}.

Recent studies on \textit{attention sink}~\citep{xiaoefficient,gu2024attention,guattention,bai_does_2024,guattention} provide key insights for analyze this phenomenon: when processing long sequences, LMs rely on stable "sink tokens" to aggregate attention and maintain semantic comprehension. For instance, \cite{xiaoefficient} demonstrates that removing these sinks causes a significant performance drop. Natural language inherently contains such sinks (e.g., key nouns, logical connectives) due to its coherent semantics, allowing LMs to learn attention patterns that aggregate information through these sinks during pre-training . However, user behavior sequences lack such explicit sink tokens: separators are semantically meaningless, and tokens within individual behaviors (e.g., item names) cannot function as cross-behavior anchors. This raises critical questions: (1) Is the absence of effective attention sinks in user behavior sequences a key reason for the performance bottleneck of LM-based CTR prediction? (2) If so, how can we introduce attention sinks to guide the model’s attention and break through this bottleneck?

To address these questions, 
we propose \textsc{CTR-Sink}: a behavior-level attention sink specifically designed for recommendation scenarios. The core idea of \textsc{CTR-Sink} is to insert tokens fused with recommendation-specific signals (e.g., temporal distance between behaviors) into user behavior sequences, serving as sink tokens to aggregate attention. To further enhance this attention aggregation capability, we propose a two-stage training strategy and a sink-specific attention mechanism. The former explicitly guides LM attention toward sink tokens
and the latter amplifies inter-sink dependencies to better capture behavioral correlations.
Experiments demonstrate that our method achieves significant improvements in AUC compared to baselines on both industrial and open-source datasets (MovieLens, Kuairec). Visualization verifies the role of \textsc{CTR-Sink} as an sink token: attention is significantly concentrated on \textsc{CTR-Sink}s, with clear and distinguishable behavioral boundaries.
This study not only promotes performance improvements in CTR prediction but also advances our understanding of attention sinks in recommendation task, bridging the gap between general LM pre-training and structured behavior sequence modeling.
To summarize, our main contributions are as follows:

\begin{itemize}
\item To the best of our knowledge, we are the first to explicitly identify semantic fragmentation in user behavior sequence modeling, analyzing its essence via attention sink theory: the lack of effective sinks causes LMs' attention to scatter from behavioral boundaries, conflicting with pre-training attention patterns .
\item We propose \textsc{CTR-Sink}, a behavior-level attention sink with recommendation-specific signals (e.g., temporal distance), complemented by a two-stage training strategy and a sink-specific attention mechanism for a complete solution.
\item We conduct extensive experiments on 1 industrial dataset and 2 open-source datasets (MovieLens, Kuairec) for both encoder (RoBERTa) and decoder (Qwen) architectures. Comparative experiments and ablation analyses verify the effectiveness of our method across different scenarios.
\end{itemize}

\section{Preliminary}
\label{preliminary}
\subsection{Problem Definition}
LM-based CTR prediction aims to estimate the probability of users clicking on an item based on text-described historical behavior sequences. Recent works typically treat CTR prediction as a binary text classification task. Here, the ground-truth labels remain consistent with traditional setups (i.e., \(y_i \in \{0, 1\}\)). These works utilize LMs to extract the dense representation \(q^{\text{text}}\) from the textual input \(x^{\text{text}}\), and then a prediction layer is applied for click estimation.
Inspired by the widely used SIM\cite{qi_search-based_2020} framework, we leverage semantic information to select the most similar behaviors, which form the final \(x^{\text{text}}\).

Formally, the list of user behaviors \( b = [b_1, b_2, b_3, \ldots, b_n] \), and the behaviors representation list \( r = [r_1, r_2, \ldots, r_n] \) is obtained through the representation model \( R \). The target behavior is \( b' \) with its representation \( r' \). We calculate the cosine similarity between \( r' \) and \( r \) respectively, and sort them in ascending order to form \( [br_1, br_2, br_3, \ldots, br_n] \). 

Finally, the input fed to the LM is 
\begin{equation}
     x^{\text{text}} = \textit{prompt} + br_1 + br_2 + \ldots + br_k ,
\end{equation}
where \( k \) is a parameter indicating the selection of \( k \) user behaviors, $+$ denotes the concatenation operation. 

\subsection{Semantic Fragmentation: Preliminary Sink Validation}
Based on the characteristics of attention sinks, we hypothesize that inserting sink-like anchor tokens into user behavior sequences may alleviate the problem of semantic fragmentation—where LM attention scatters across irrelevant tokens and fails to focus on behavior boundaries or inter-behavior relationships (as visualized in Figure~\ref{fig:combined}). In encoder models such as BERT, the [CLS] token serves as an anchor for semantic aggregation during pre-training (a natural attention sink). Therefore, we initially attempt to insert [CLS] tokens into user behavior sequences to observe their effect on guiding attention. Specifically, we modify the input format as follows:
\begin{equation}
x^{\text{text}} = \textit{prompt} + br_1 + [\text{CLS}] + \ldots + [\text{CLS}] + br_k.
\end{equation}
Experimental results are presented in Table~\ref{table:roberta}. We adopt Area Under the ROC Curve(AUC)\cite{bradley1997use} as the evaluation metric. 

\begin{table}[h!]
\centering
\caption{Performance of [CLS] Tokens in Mitigating Semantic Fragmentation. 
}
\begin{tabular}{l c c c c} 
\toprule
  & Industry & Movielens & Kuairec \\ \midrule
LM-CTR & 0.7764 & 0.7808 & 0.8133 &  \\ \midrule
LM-CTR w [CLS] & 0.7774 & 0.7820 & 0.8141 &  \\ 
\bottomrule

\end{tabular}

\label{table:roberta}
\end{table}

Experimental results show that:
1) On the industrial dataset, the model performance improves by 0.1\%; on MovieLens and Kuairec, the improvements are 0.12\% and 0.08\%, respectively. \textbf{\textit{It is worth noting that an AUC increase of 0.001(0.1\%) can be considered a significant improvement in CTR prediction\cite{wang_bert4ctr_2023,zhang_collm_2025,li_ctrl_2023,wang_flip_2024}.}}
2) Attention visualization (Appendix~\ref{appendix:cls_visualization}) indicates that [CLS] tokens attract more attention, enabling the model to start aggregating information by behavior units, which preliminarily verifies the hypothesis that "implanting sink anchors can alleviate semantic fragmentation."

However, as a general aggregation token, [CLS] does not incorporate recommendation-specific information (e.g., behavior temporality), leading to limited effectiveness. This inspires us to design a more tailored attention sink for recommendation tasks—\textit{CTR-Sink}.
\section{\textsc{CTR-Sink}}
In the Preliminary section, we verified the core idea that "implanting attention sinks can alleviate semantic fragmentation in user behavior sequences" through preliminary experiments with [CLS] tokens. However, this exploration still has significant limitations: as a general semantic aggregation token, [CLS] does not incorporate signals specific to recommendation scenarios (e.g., behavior temporality) and exhibits poor adaptability to decoder architectures. 

Based on this, we need to address the particularities of language model attention regulation in recommendation scenarios and systematically solve the following three progressive problems to improve the design logic of the \textsc{CTR-Sink} method:

\paragraph{\textbf{Section~\ref{sec:landmark_design}} Generic [CLS] tokens, lacking recommendation-domain signals (e.g., behavior temporality, importance), struggle to serve as efficient attention sinks. Can we design a special token carrying recommendation-specific external information (e.g., temporal distance between a behavior and the target item) to make them more scenario-adaptive attention sinks, thereby precisely guiding LM attention to focus on behavior boundaries?}

\paragraph{\textbf{Section~\ref{sec:two_stage_training}} Decoder architectures (e.g., Qwen) naturally lack mechanisms for capturing attention to sinks due to the absence of learning processes involving [CLS]-like sink tokens during pre-training, which weakens the effectiveness of CTR-Sink. How can we optimize training strategies to enhance the ability of decoder models to attend to CTR-Sinks?}

\paragraph{\textbf{Section~\ref{sec:attention_bias}} Recommendation scenarios require not only identifying behavior boundaries but also capturing inter-behavior correlations (e.g., the underlying logic of "watching Movie A tends to lead to watching Movie B"). After sink tokens achieve attention focusing, how can we strengthen attention connections between sinks to model the overall semantics of behavior sequences?}

\begin{figure*}[htbp]
    \includegraphics[width=0.7\textwidth]{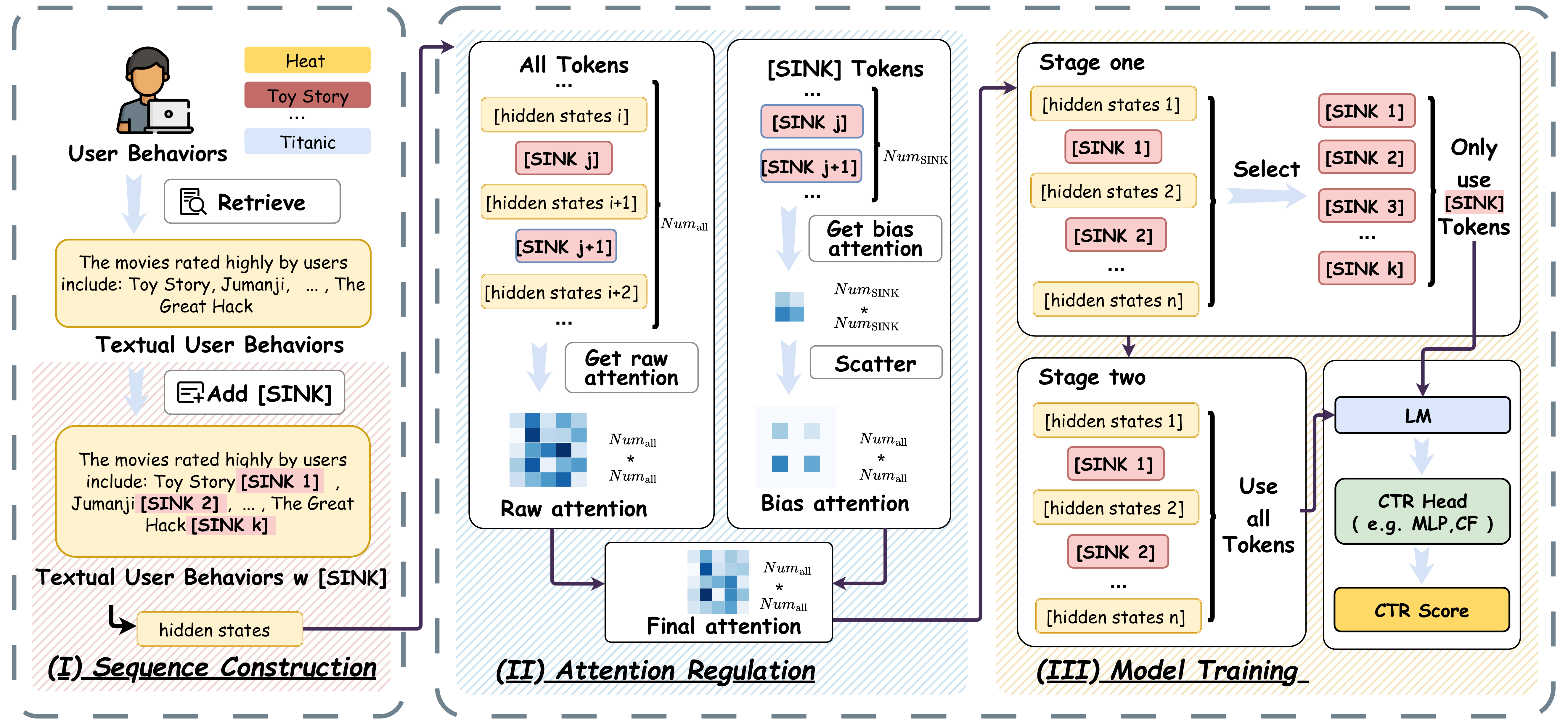}
    \caption{An intuitive illustration of the \textsc{CTR-Sink}. Specifically, \textit{(I). Sequence Construction} displays insertion of [\textit{SINK}] tokens into user behavior sequences.
    \textit{(II) Attention Regulation} illustrates how the sink-specific mechanism strengthens inter-behavior association modeling.
    \textit{(III) Model Training} shows two-stage training: first boosting sink attention, then overall prediction optimization.
}
    \label{main figure}
\end{figure*}

\subsection{Can Recommendation-Specific Information Enhance Attention Sinks?}
\label{sec:landmark_design}
\begin{table*}[htbp]
\centering
\caption{Performance impact of adding [\textit{SINK}] and two-stage training on RoBERTa and Qwen.}
\begin{tabular}{lcccccc}
\toprule
\multirow{2}{*}{\textbf{Method}} & \multicolumn{3}{c}{\textbf{RoBERTa}} & \multicolumn{3}{c}{\textbf{Qwen}}\\
                                    \cmidrule(lr){2-7}  
& \textbf{Industry} & \textbf{MovieLens} & \textbf{KuaiRec} & \textbf{Industry} & \textbf{MovieLens} & \textbf{KuaiRec} \\
\midrule
Original  & 0.7764 & 0.7808 & 0.8133 & 0.7885 & 0.8177 & 0.8156 \\
w [\textit{SINK}]  & 0.7784 & 0.7836 & 0.8175 & 0.7891 & 0.8181 & 0.8172 \\
w Two-stage training & 0.7791 & 0.7840 & 0.8181 & 0.7902 & 0.8193 & 0.8189 \\
\bottomrule
\end{tabular}

\label{table:landmark_and_twostage}
\end{table*}

Generic [CLS] tokens struggle to serve as efficient attention sinks due to their lack of recommendation-domain signals (e.g., behavior temporality, importance). To address this, we design a \textbf{information-enhanced special token} that incorporate recommendation-specific external information and validate their effectiveness as attention sinks.

The core design of information-enhanced sinks involves inserting special tokens fused with recommendation-domain signals into user behavior sequences. Specifically:
Given the original sequence \([b_1, b_2, b_3, \ldots, b_n] \) and the retrieved sequence \([br_1, br_2, br_3, \ldots, br_n] \), we modify the input as 
\begin{equation}
x^{\text{text}} = \textit{prompt} + br_1 + [\text{\textit{SINK}}_1] + br_2 + [\text{\textit{SINK}}_2] + \ldots + br_k.
\end{equation}
\textbf{In this paper, [\textit{SINK}] denotes the information-enhanced special token}. Notably, adding one [\textit{SINK}] per behavior (typically approximately 10-15 tokens) incurs negligible overhead ($<10\%$ length increase). 
Specifically, we construct the [\textit{SINK}] as follows:
\begin{equation}
[\text{\textit{SINK}}_i] = \text{\textit{MLP}}(\text{\textit{Embed}}(r_i - i)),
\end{equation}
where \( r_i - i \) represents the temporal distance between \( br_i \) and the target behavior. The retrieval index \(r_i\) is produced by a pre-trained Contriever, whose parameters are frozen during training and thus introduce no information leakage from the main LM. This design is rooted in the recommendation-domain consensus that recent behaviors have a more significant impact on users' current preferences \cite{hou2023deep,liu2023deep,feng2024context,feng_long-sequence_2024}. We select temporal signals here due to their intuitive effectiveness and widespread availability in most datasets. Furthermore, we conduct ablation studies in Section~\ref{sec:Ablation of External Information} to validate the efficacy of alternative signals. 

The experimental results show that:
1) As shown in Table \ref{table:landmark_and_twostage}, with only a small increase in the number of tokens, the improvement on RoBERTa is extremely significant, achieving a 0.2\% improvement on the industrial dataset, 0.28\% on MovieLens, and a remarkable 0.42\% improvement on Kuairec.
2) Meanwhile, the attention heatmap (Figure \ref{fig:heap_roberta}) further confirms that a large amount of attention in RoBERTa is concentrated on the [\textit{SINK}] tokens, indicating that \textbf{[\textit{SINK}], as a special token carrying external information, effectively addresses the problem of attention dispersion and bridges the gap between the special token and the sink token.}
3) \textit{However, we noticed that the improvement on Qwen is not as significant}, especially on the MovieLens dataset where there is only a marginal 0.04\% improvement. This stagnation suggests that information alone is insufficient: without the explicit attention pattern, the model fails to perceive the injected temporal signals. Thus, the sink mechanism is a prerequisite carrier—it must be actively awakened to enable the utilization of such external information.

\subsection{How to Boost Decoder Attention on \textsc{CTR-Sink}?}
\label{sec:two_stage_training}
\begin{figure}[htbp]
  \centering
  \subcaptionbox{RoBERTa.
  \label{fig:heap_roberta}}{%
    \includegraphics[width=0.23\textwidth]{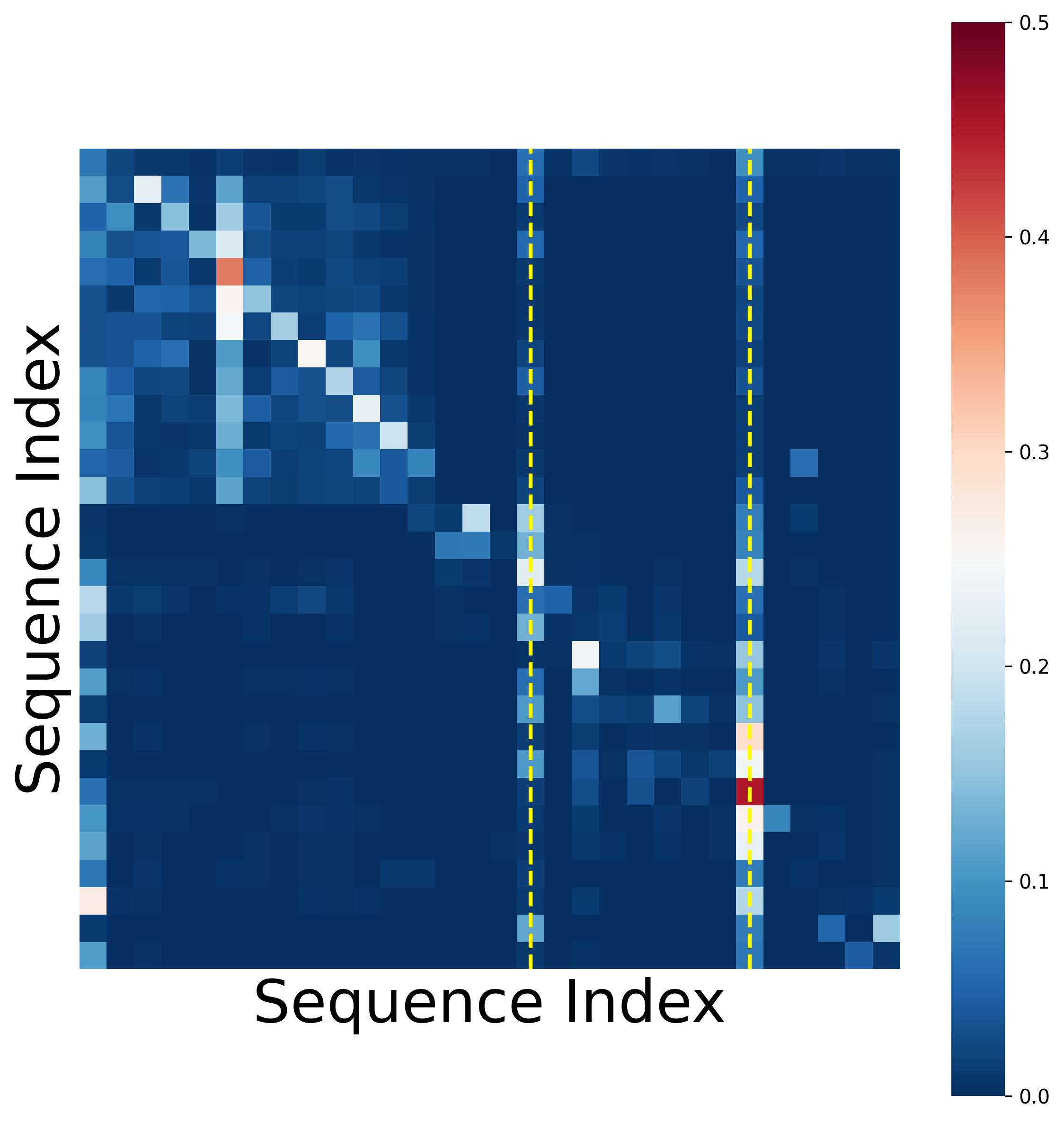}
  }
  \hfill
  \subcaptionbox{Qwen.
  \label{fig:heap_qwen}}{%
    \includegraphics[width=0.23\textwidth]{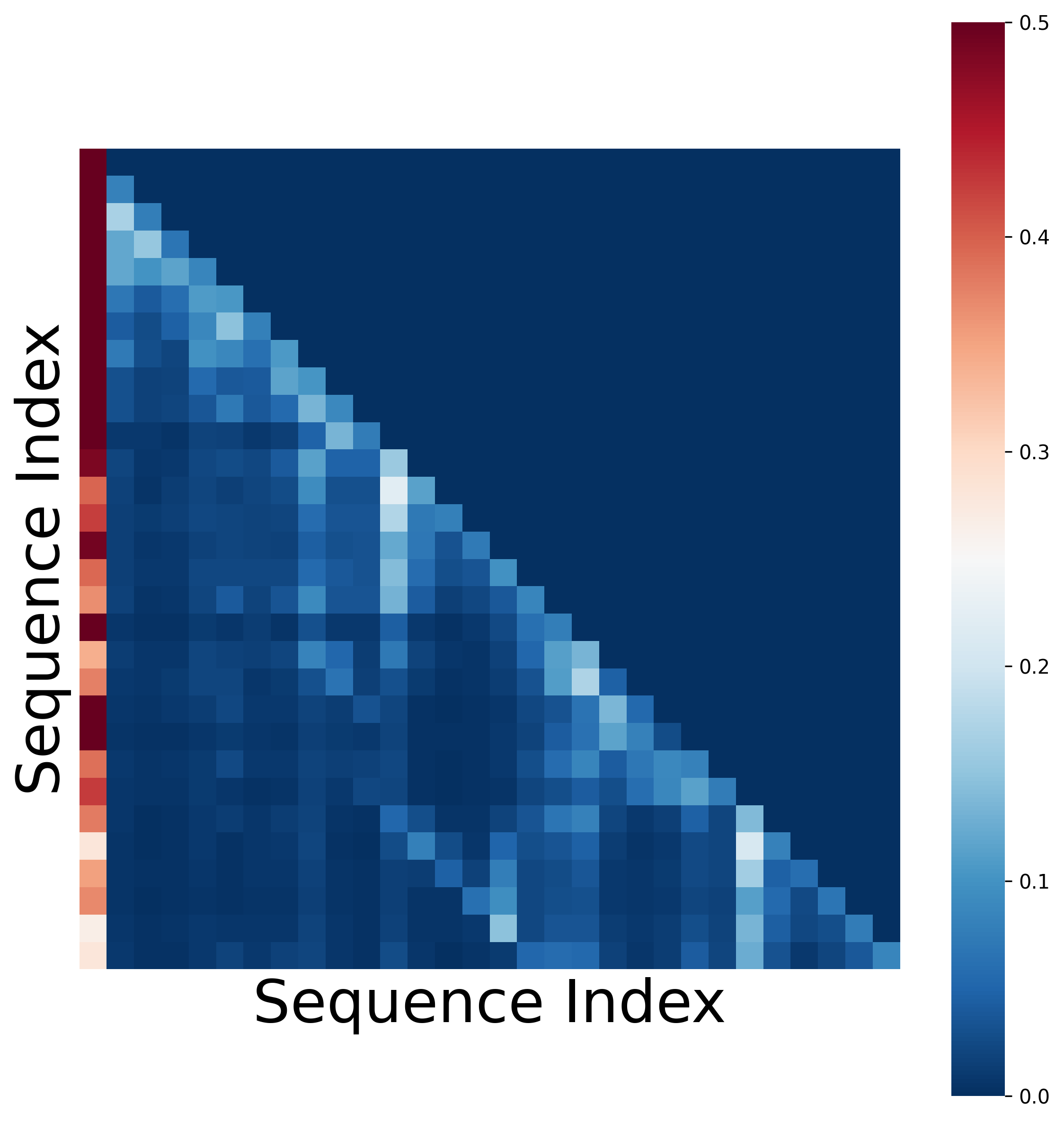}
  }

  \caption{Attention Heatmaps of RoBERTa and Qwen with [\textit{SINK}]. In (a) RoBERTa, attention clearly aggregates at the positions of [\textit{SINK}] tokens marked by the yellow line, whereas in (b) Qwen, attention largely concentrates on the first token.}
  \label{fig:attention_heatmap}
  \vspace{-5mm}
\end{figure}

We argue that the limited effectiveness of directly adding special tokens([\textit{SINK}]) to decoder architectures like Qwen stems from two main reasons: 
(1) Decoder architectures inherently lack the ability to attend to attention sinks, as they are not pre-trained with sink tokens such as [CLS]. 
(2) Additionally, existing works have pointed out that decoder models, as illustrated in Figure \ref{fig:heap_qwen}, tend to concentrate significant attention on the first token\cite{xiaoefficient,guattention}, which further weakens the aggregating effect of [\textit{SINK}] as attention sinks. 

To address this, we design a \textit{two-stage training strategy} to enhance their perception of sinks: 
Specifically, during normal training, the hidden layers of all tokens output by the LM are fed into the CTR prediction module, with the objective function of minimizing 
\begin{equation}
\mathcal{L} = \frac{1}{|\mathcal{D}|} \sum_{u \in U, i \in I} l\bigl(Y, f(LM(All\ tokens))\bigr),
\end{equation} 
where \( l(\cdot) \) denotes the loss function (e.g., cross-entropy loss), \( |\mathcal{D}| \) is the total number of training samples, \( Y \) indicates the binary click label (1 for click, 0 for non-click), and \( f(\cdot, \cdot) \) represents the CTR prediction function of our model. 

Prior to this, we add a \textit{first-stage training}, where only the landmark tokens are input into the CTR prediction module for prediction. The training objective function here is to minimize 

\begin{equation}
\mathcal{L} = \frac{1}{|\mathcal{D}|} \sum_{u \in U, i \in I} l\bigl(Y, f(LM([\textit{SINK}]\ tokens))\bigr).
\end{equation} 
Its core role is to force the model to predict click-through rates solely through [\textit{SINK}] tokens, guiding attention to gather toward the sinks. 
Subsequently, conventional training is performed to optimize the final prediction using full-sequence information while retaining the attention to sinks.

Experimental results in Table \ref{table:landmark_and_twostage} show that after two-stage training, the AUC of RoBERTa on MovieLens increases to 0.7840, and on Kuairec to 0.8181; Qwen achieves more significant improvements, with the AUC reaching 0.8193 on MovieLens and 0.8189 on Kuairec.

\begin{figure}[htbp]
  \centering
  \subcaptionbox{Attention proportion focused on [\textit{SINK}] tokens across layers.\label{fig:Attention_Focused_on_Landmarks}}{%
    \includegraphics[width=0.34\textwidth]{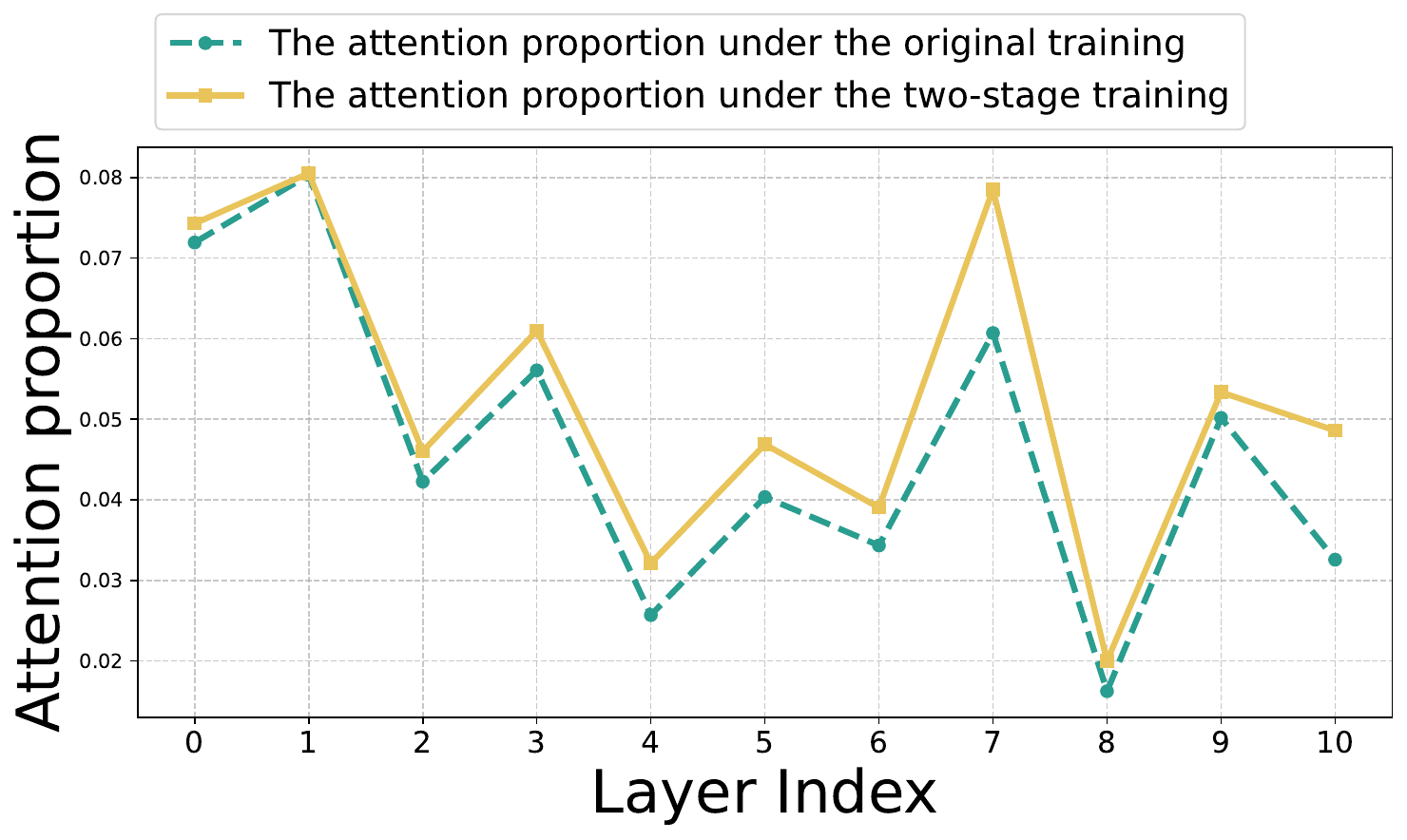}
  }
  \subcaptionbox{Attention proportion between [\textit{SINK}] tokens across layers.\label{fig:Attention_between_Landmarks}}{%
    \includegraphics[width=0.34\textwidth]{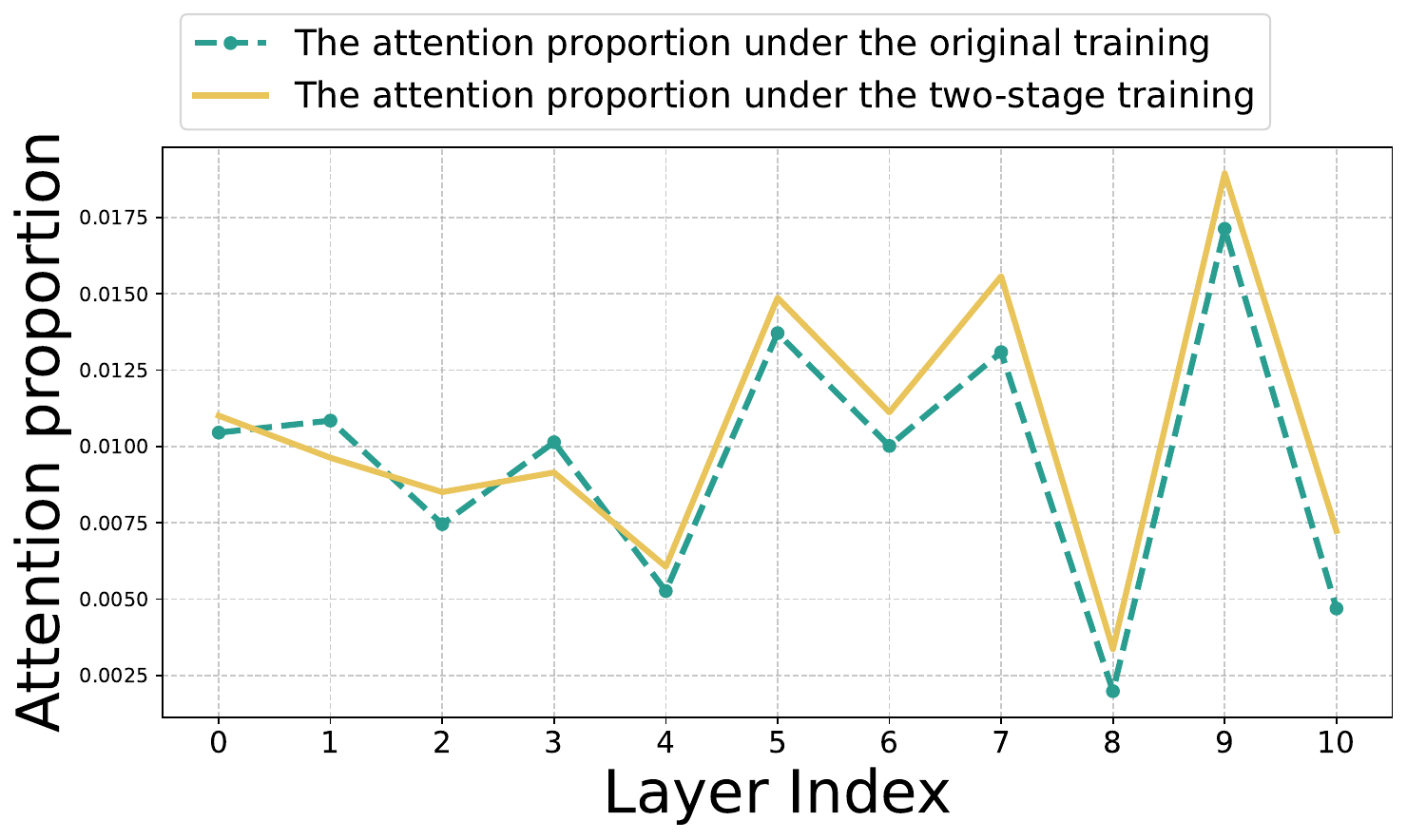}
  }
  
  \caption{Attention proportion on [\textit{SINK}] tokens. (a) shows two-stage training increases attention on [\textit{SINK}] tokens. (b) further shows, in deeper layers (above layer 4), inter-[\textit{SINK}] token attention rises, meaning one [\textit{SINK}] token focuses more on others.
}
  \label{fig:attention_proportion}
  \vspace{-5mm}
\end{figure}

To more intuitively demonstrate the degree of such attention aggregation, Figure~\ref{fig:Attention_Focused_on_Landmarks} plots the attention proportion focused on [\textit{SINK}] tokens across each layer of the model. Specifically, 

\begin{equation}
P_{f} = \frac{\sum_{i} \sum_{j \in L} A_{i,j}}{\sum_{i,j} A_{i,j}},
\end{equation}
where $L$ denotes the set of [\textit{SINK}].
The results in Figure~\ref{fig:Attention_Focused_on_Landmarks} indicate that the proportion of the model's attention directed toward [\textit{SINK}] has significantly increased, effectively mitigating the sink perception deficiency of the decoder architecture.

Thus, we can draw two conclusions: 
1) \textbf{The aggregation of attention on attention sinks is highly beneficial for modeling text-based user historical behaviors.} 
2) \textbf{This two-stage training approach can enhance the model's ability to aggregate information.}

\subsection{How to Strengthen Sink Associations for Deep Behavioral Semantics?}
\label{sec:attention_bias}

In recommendation scenarios, it is usually necessary to focus on capturing inter-behavior correlations (e.g., "watching Movie A tends to lead to watching Movie B"), but existing methods insufficiently model the associations between attention sinks. For intuitive demonstration, Figure \ref{fig:Attention_between_Landmarks} plots the attention proportion between [\textit{SINK}] tokens. Specifically, 
\begin{equation}
P_{b} = \frac{\sum_{i \in L} \sum_{j \in L} A_{i,j}}{\sum_{i,j} A_{i,j}},
\end{equation}
where $L$ denotes the set of [\textit{SINK}]. This attention proportion indicates the extent to which [\textit{SINK}] tokens attend to other [\textit{SINK}] tokens.
We find that after two-stage training, the attention proportion between [\textit{SINK}] does not significantly increase or decrease in shallow layers (below layer 4), but significantly increases in deeper layers (above layer 4). This finding is consistent with the observation that information of anchor tokens aggregates in deeper layers\cite{wang_label_2023}, inspiring us to enhance the regulation of attention between [\textit{SINK}]s.

To this end, we design a \textit{sink-specific attention enhancement mechanism}. On the basis of the model's original attention calculation, this mechanism adds a bias generated by an independent self-attention mechanism to the attention weights between each pair of [\textit{SINK}] tokens, as follows:
\begin{table}[htbp]
\centering
\caption{Ablation of sink-specific inter-sink attention enhancement.}
\begin{tabular}{lccc}
\toprule
& \textbf{Industry} & \textbf{MovieLens} & \textbf{KuaiRec} \\
\midrule
w/o inter-sink attention & 0.7791 & 0.7840 & 0.8181 \\
w inter-sink attention & 0.7810 & 0.7844 & 0.8192 \\
\bottomrule
\end{tabular}
\label{table:intersink_attention}
\end{table}

1. \textit{Original Attention Calculation}: For a sequence $\mathbf{X} = [x_1, x_2, \ldots, x_n]$ containing all tokens (a total of $n$ tokens), the original attention weight matrix is first calculated:  
   \begin{equation}
\mathbf{Attn}_{\text{raw}}{(n \times n)} =  \frac{\mathbf{Q} \cdot \mathbf{K}^\top}{\sqrt{d_{\text{qkv}}}} ,
\end{equation}
   where $\mathbf{Q}$ and $\mathbf{K}$ are the query and key vectors of all tokens, respectively, and $\mathbf{Attn_{raw}}$ has a dimension of $n \times n$.

2. \textit{Bias Matrix Generation and Scattering}: For the set of [\textit{SINK}] tokens $\mathcal{L} = [l_1, l_2, \ldots, l_K]$ (a total of $k$ tokens, $k \ll n$), their bias matrix is calculated through an independent self-attention mechanism:  
   \begin{equation}
\mathbf{Attn}_{\text{bias}}{(k \times k)} =   \frac{\mathbf{Q}_{\text{bias}}' \cdot \mathbf{K}_{\text{bias}}'^\top}{\sqrt{d_{\text{bias}}}} ,
\end{equation}
   where $\mathbf{Q}_{\text{bias}}' = \mathbf{X}_\mathcal{L} \cdot W_Q''$ and $\mathbf{K}_{\text{bias}}' = \mathbf{X}_\mathcal{L} \cdot W_K''$ ($\boldsymbol{X}_\mathcal{L}$ denotes the hidden states of [\textit{SINK}] tokens $\mathcal{L}$, extracted from the sequence representation after prior encoding layers, $W_Q''$ and $W_K''$ are linear transformation parameters independent of the original attention). Subsequently, the $k \times k$ $\mathbf{attention_{bias}}$ is "scattered" into an $n \times n$ matrix according to the positions of [\textit{SINK}] tokens in the original sequence (only the positions corresponding to [\textit{SINK}] retain bias values, and the rest are 0).

3. \textit{Fused Attention Weights}:  
   \begin{equation}
\begin{aligned}
\mathbf{Attn}_{\text{final}}{(n \times n)} &= \text{\textit{softmax}}\Big( \mathbf{Attn}_{\text{raw}}{(n \times n)} \\
& \quad + \mathbf{Attn}_{\text{bias}}{(n \times n)} \Big)
\end{aligned}.
\end{equation}
   The final $n \times n$ attention weight matrix is obtained, which not only retains the global distribution of the original attention but also strengthens the association weights between [\textit{SINK}] tokens.

The overall workflow is illustrated in Figure~\ref{main figure}. The direct comparison in Table~\ref{table:intersink_attention} further confirms the effectiveness of this module: explicitly strengthening inter-sink attention improves the model across all three datasets, showing that inter-behavior correlations are better captured when [\textit{SINK}] tokens are allowed to interact more strongly.

\section{Experimental Setup}
\subsection{Datasets}
We conduct experiments on one real-world industrial dataset and two open-source datasets (MovieLens\cite{harper2015movielens} and KuaiRec\cite{gao2022kuairec}). Among them, the industrial dataset and KuaiRec are Chinese datasets, while MovieLens is an English dataset. The raw industrial logs contain approximately 100 million exposures, 700 thousand users, and 200 thousand items, with a click ratio of 33.44\%. For KuaiRec, we filter out samples lacking textual descriptions and define samples with a completion rate exceeding 100\% as positive, and the rest as negative. For MovieLens, to further improve data quality, we select users with at least 20 records and use data from the past decade, with the binary classification threshold set to 3. All datasets are split into training, validation, and test sets by date, following an 8:1:1 ratio. Detailed statistics of the datasets after filtering are provided in Appendix ~\ref{appendix:dataset}.

\subsection{Evaluation Metrics}
To evaluate the performance of CTR prediction methods, we adopt Area Under the ROC Curve(AUC)\cite{bradley1997use} as the evaluation metric.
\textit{Notably, the AUC improvements of 0.2\%-0.5\% are statistically significant in CTR prediction, where an increment of 0.001(0.1\%) is regarded as non-trivial\cite{wang_bert4ctr_2023,zhang_collm_2025,li_ctrl_2023,wang_flip_2024}.}

\subsection{Baseline Models}
\label{sec:4.3}
For traditional CTR models, our baselines include four ID-based DNN models: DeepFM \cite{guo2017deepfm}, DIN \cite{zhou2018deep}, AutoInt \cite{song2019autoint}, and DCN-V2 \cite{wang2021dcn}.
For LM-based approaches, since our CTR-Sink method is model-agnostic, we adopt a mainstream LM-CTR framework\cite{bao2023tallrec,li_ctrl_2023,zhang_collm_2025,chang_singleton_2025,geng_breaking_2024} as the baseline. We note that recent works such as SCTR \cite{chang_singleton_2025} and CoLLM \cite{zhang_collm_2025} typically leverage LMs to directly generate 'yes/no' tokens for prediction. In contrast, our LM-CTR extracts dense representations from the LM's final layer and feeds them into a specialized CTR prediction head. Despite this divergence in output paradigms, the core mechanism of encoding textual behavior sequences remains fundamentally consistent. To demonstrate the universality of our method across architectures, we implement LM-CTR on both RoBERTa (encoder) and Qwen (decoder). Furthermore, we include CoLLM~\cite{zhang_collm_2025} and FLIP-PLM~\cite{wang_flip_2024} to compare with representative LM-based methods that inject collaborative or ID information into language models.

\subsection{Implementation Details}
We adopt AdamW as the optimizer. For the experiments involving our backbone models (RoBERTa\cite{liu2019roberta} and Qwen2-0.5B\cite{team2024qwen2}), across all datasets, we set the batch size to 64 and the learning rate to 1e-4 for RoBERTa, and set the batch size to 16 and the learning rate to 1e-5 for Qwen2. We select a warm ratio of 0.05 and set the number of epochs to 3 for non-two-stage training. When generating [\textit{SINK}] tokens, we set the dimension of the embedding layer to 128 for RoBERTa and 256 for Qwen(We analyze the sensitivity of this parameter in Appendix~\ref{appendix:dim}), and the subsequent MLP is a linear layer whose output matches the hidden layer dimension of the corresponding LM. In the sink-specific attention enhancement mechanism, we use an attention layer that matches the hidden layer dimension and the number of attention heads of the corresponding LM, and set the dropout to 0.1. All tests run on 8 A100 GPUs. 

\section{Experimental Results}

 In this section, we answer the following research questions:
\begin{enumerate}
    \item[RQ1] How does \textsc{CTR-Sink} perform compared to the baseline?
    \item[RQ2] Can the external information of [\textit{SINK}] be replaced with other types of information?
    \item[RQ3] Will the performance of \textsc{CTR-Sink} degrade as the number of behaviors increases?
    \item[RQ4] Is the improvement in the two-stage training due to the increase in epochs?
\end{enumerate}

\subsection{Overall Performance(RQ1)}
\label{sec:5.1 overall result}
\begin{table}[htbp]
\centering
\caption{Overall performance. $\star$Indicates the improvements are statistically significant with p < 0.005.}
\begin{tabular}{lccc}
\toprule
\multirow{2}{*}{\textbf{Method}} & \multicolumn{1}{c}{\textbf{Industrial}} & \multicolumn{2}{c}{\textbf{Open source }}\\
                                    \cmidrule(lr){2-4}  
& \textbf{Industry} & \textbf{MovieLens} & \textbf{KuaiRec} \\
\midrule
DeepFM  & 0.7678 & 0.7803 & 0.8072 \\
DIN  & 0.7712 & 0.7785 & 0.8086 \\
AutoInt  & 0.7735 & 0.7812 & 0.8064 \\
DCN-V2  & 0.7719 & 0.7806 & 0.8065 \\

\midrule
\multicolumn{4}{@{\hspace{0.22cm}}l}{\textbf{\textit{RoBERTa}}} \\
\hdashline
CoLLM & 0.7695 & 0.7795 & 0.8097 \\
CoLLM w \textsc{CTR-Sink} & \textbf{0.7787}$\star$ & \textbf{0.7829}$\star$ & \textbf{0.8155}$\star$ \\
\hdashline
FLIP-PLM & 0.7792 & 0.7820 & 0.8137 \\
FLIP-PLM w \textsc{CTR-Sink} & \textbf{0.7823}$\star$ & \textbf{0.7852}$\star$ & \textbf{0.8195}$\star$ \\
\hdashline
LM-CTR  & 0.7764 & 0.7808 & 0.8133 \\
LM-CTR w \textsc{CTR-Sink} & \textbf{0.7810}$\star$ & \textbf{0.7844}$\star$ & \textbf{0.8192}$\star$ \\
\midrule
\multicolumn{4}{@{\hspace{0.22cm}}l}{\textbf{\textit{Qwen}}} \\
\hdashline
CoLLM & 0.7888 & 0.8165 & 0.8092 \\
CoLLM w \textsc{CTR-Sink} & \textbf{0.7924}$\star$ & \textbf{0.8195}$\star$ & \textbf{0.8125}$\star$ \\
\hdashline
LM-CTR  & 0.7885 & 0.8177 & 0.8156 \\
LM-CTR w \textsc{CTR-Sink} & \textbf{0.7919}$\star$ & \textbf{0.8203}$\star$ & \textbf{0.8198}$\star$ \\
\bottomrule
\end{tabular}

\label{table:main_result}
\end{table}

The experimental results in Table~\ref{table:main_result} demonstrate that \textsc{CTR-Sink} consistently outperforms all baseline models across diverse datasets and model architectures, directly addressing RQ1. For encoder-based models (RoBERTa), integrating \textsc{CTR-Sink} yields significant AUC improvements: 0.46\% on the industrial dataset, 0.36\% on MovieLens, and 0.59\% on KuaiRec. Similarly, decoder-based models (Qwen) achieve improvements of 0.34\%, 0.26\%, and 0.42\% on the three datasets, respectively.

It is also worth noting that \textsc{CTR-Sink} improves both CoLLM and FLIP-PLM, which respectively represent generative collaborative alignment and feature-level ID--PLM alignment. This indicates that these alignment mechanisms and our behavior-level attention regulation are orthogonal: they enhance how collaborative or ID information is injected into LMs, while \textsc{CTR-Sink} further alleviates semantic fragmentation in concatenated behavior sequences.

 The consistent performance gains across diverse datasets (Chinese/English, industrial/open-source) and architectural types (encoder/decoder) demonstrate that \textsc{CTR-Sink} effectively mitigates semantic fragmentation by functioning as behavior-level attention sinks, thus confirming its generalizability. Compared to traditional ID-based DNN models, the proposed approach retains the semantic modeling advantages of language models while overcoming their inherent limitations in processing unstructured behavior sequences.

To further verify robustness under stronger decoder backbones, we additionally conduct MovieLens experiments with Qwen2-7B and Qwen3-8B, and report the results in Appendix~\ref{appendix:large_qwen}. These supplementary results show that \textsc{CTR-Sink} remains effective at larger model scales, supporting the robustness of behavior-level attention regulation.

\begin{figure}[htbp]
    \includegraphics[width=0.34\textwidth]{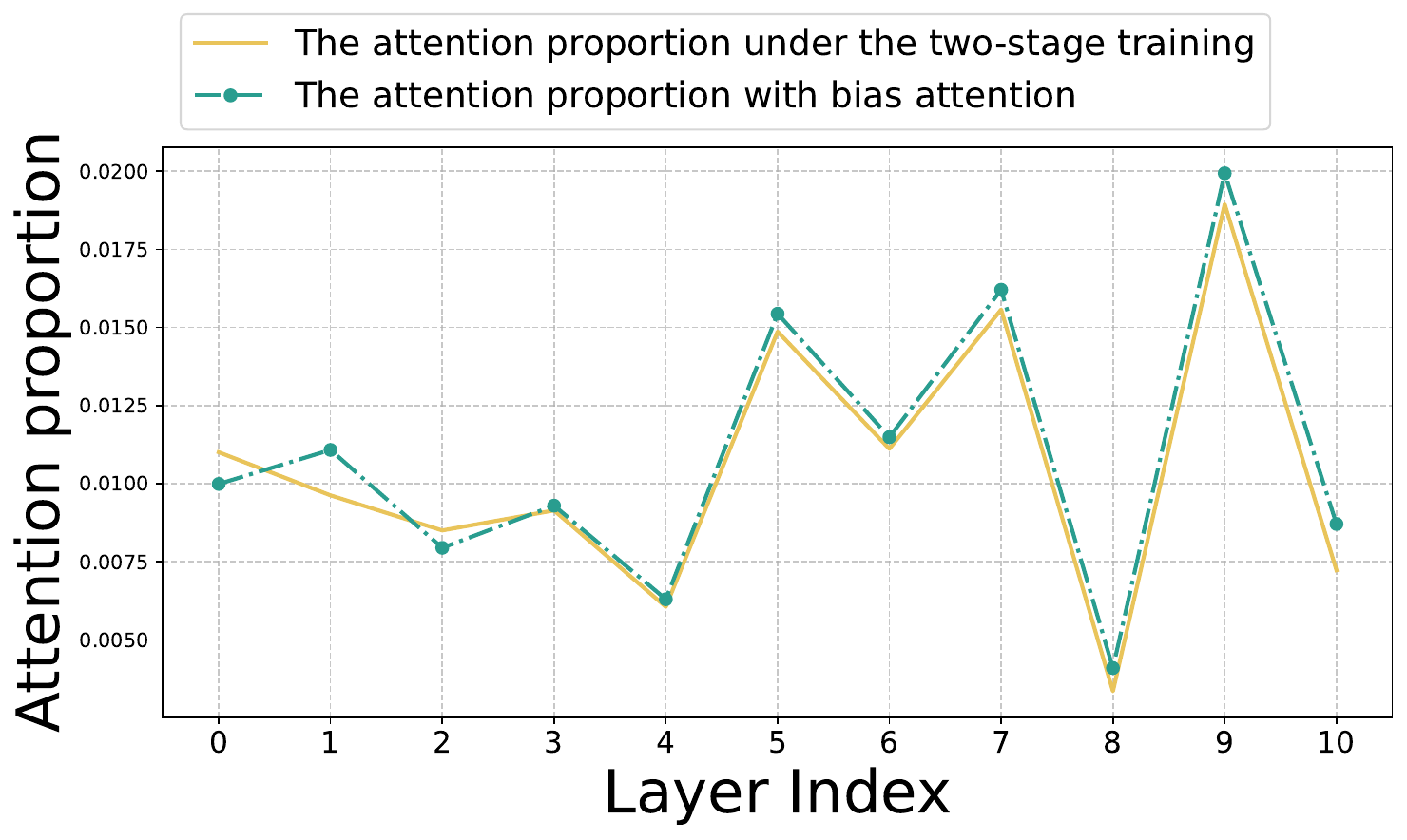}
    \caption{Attention proportion between [\textit{SINK}] tokens
across layers after adding sink-specific attention enhancement mechanism.}
    \label{fig:attention after bias}
\end{figure}

Figure~\ref{fig:attention after bias} shows that the sink-specific attention enhancement mechanism increases the proportion of attention between [\textit{SINK}] tokens in deeper layers. This observation confirms that strengthening inter-sink dependencies facilitates more effective capture of inter-behavioral correlations, thereby improving the model's ability to model complex behavioral semantics—directly supporting the hypothesis that \textbf{attention interaction between sinks is critical for textual user behavior modeling}.

\subsection{Ablation Studies}

\subsubsection{Ablation of External Information(RQ2)}
\label{sec:Ablation of External Information}
To address RQ2 regarding whether [\textit{SINK}] external information can be replaced by other types, we conduct ablation studies on two types of signals (behavior temporality vs. semantic similarity) and random controls, with results presented in Table~\ref{table:ablation_extern_info}. "Use semantic similarity as external information" refers to generating [\textit{SINK}] by replacing the original temporal distance with semantic similarity while maintaining the real chronological order of user behavior sequences; "w Random [\textit{SINK}]" means that the additional information carried by [\textit{SINK}] is randomly assigned, so as to verify the necessity of recommendation-specific signals in guiding the model to pay attention to different sinks.
\begin{table}[htbp]
\centering
\caption{Ablation of External Information.}
\begin{tabular}{lccc}
\toprule
& \textbf{Industry} & \textbf{MovieLens} & \textbf{KuaiRec} \\
\midrule
\multicolumn{4}{@{\hspace{0.22cm}}l}{\textbf{\textit{Use behavior
temporality as external information}}} \\
\hdashline
Original  & 0.7764 & 0.7816 & 0.8133 \\
w Random [\textit{SINK}] & 0.7768 & 0.7813 & 0.8136 \\
w [\textit{SINK}] & 0.7810 & 0.7844 & 0.8192 \\
\midrule
\multicolumn{4}{@{\hspace{0.22cm}}l}{\textbf{\textit{Use semantic similarity as external information}}} \\
\hdashline
Original  & 0.7718 & 0.7762 & 0.8120 \\
w Random [\textit{SINK}] & 0.7723 & 0.7764 & 0.8127 \\
w [\textit{SINK}] & 0.7764 & 0.7790 & 0.8169 \\

\bottomrule
\end{tabular}

\label{table:ablation_extern_info}
\end{table}

The experimental results show that both [\textit{SINK}] based on temporal information and semantic similarity information can effectively guide the model's attention, significantly outperforming the original model without [\textit{SINK}] and far surpassing the benchmark with random information. Under the setting of temporal information, the model's AUC on the industrial dataset, MovieLens, and KuaiRec increases by 0.46\%, 0.28\%, and 0.59\% respectively. Although the effect of semantic similarity information is slightly inferior to that of temporal information, it still achieves AUC improvements of 0.46\%, 0.28\%, and 0.49\% on the industrial dataset, MovieLens, and KuaiRec respectively, indicating that as a recommendation-related signal (reflecting semantic associations between behaviors), it can also provide effective guidance for sinks and proves the value of non-temporal specific signals.

In contrast, the improvement of w Random [\textit{SINK}] is negligible (e.g., only 0.04\% on the industrial dataset, and even a 0.03\% decrease on MovieLens), which directly indicates that random information cannot make the model form differential attention to different sinks, further highlighting the key role of recommendation-specific signals such as temporality and semantic similarity—they can inject interpretable behavioral association clues into [\textit{SINK}], enabling the model's attention to gather at key behavioral boundaries, thereby breaking through the bottleneck of semantic fragmentation. 
Additional comparisons with frequency and multi-signal variants are provided in Appendix~\ref{appendix:external_signal_variants}; frequency is only evaluated on datasets where it is available.

\subsubsection{Impact of Number of Behaviors on Performance(RQ3)}
\begin{table}[h!]
\centering

\caption{Performance comparison of different number of behaviors on MovieLens.}
\begin{tabular}{l c c c c c} 
\toprule
Number of behaviors   & 20 & 30 & 40 & 50\\ \midrule
Original  & 0.7658 & 0.7763 & 0.7814 & 0.7816     \\ \midrule
w \textsc{CTR-Sink}   & 0.7674 & 0.7782 & 0.7828 & 0.7844  \\  \midrule
Improvement  & 0.0016 & 0.0019 & 0.0014 & 0.0028 \\
\bottomrule

\end{tabular}

\label{table:ablation_length}
\end{table}
To address RQ3 (whether \textsc{CTR-Sink}'s performance degrades with increasing sequence length), we evaluate the model's behavior on the MovieLens dataset using RoBERTa, varying the number of user behaviors (20, 30, 40, 50). The results in Table~\ref{table:ablation_length} reveal critical insights into the model's adaptability to long sequences.

The original method exhibits a clear performance saturation trend with longer behavior sequences. Its AUC improves marginally from 0.7658 (20 behaviors) to 0.7814 (40 behaviors), with a mere 0.02\% gain between 40 and 50 behaviors (stagnating at 0.7816), indicating a significant bottleneck in modeling extended behavioral contexts. This phenomenon aligns with our earlier analysis of semantic fragmentation: as sequences lengthen, the model's attention scatters across irrelevant tokens, failing to effectively aggregate information at behavioral boundaries or capture inter-behavior correlations.

In contrast, the model integrated with \textsc{CTR-Sink} maintains steady performance gains across all sequence lengths. Specifically, it achieves AUC improvements of 0.16\% (20 behaviors), 0.19\% (30 behaviors), 0.14\% (40 behaviors), and 0.28\% (50 behaviors) compared to the original method. Notably, the largest gain (0.28\%) occurs at the longest sequence length (50 behaviors), demonstrating \textsc{CTR-Sink}'s enhanced capability to handle long behavioral sequences.

This discrepancy confirms that \textsc{CTR-Sink}'s behavior-level attention sinks effectively mitigate the attention dispersion issue in long sequences. By anchoring attention at behavioral boundaries and strengthening inter-sink dependencies, the method enables robust modeling of extended user behaviors, overcoming the inherent limitations of LMs in processing unstructured behavioral data. These results further validate the generalizability of \textsc{CTR-Sink}, particularly its superiority in scenarios requiring long-sequence user behavior modeling.

\subsubsection{Ablation of Epoch Count: Validating Two-Stage Training Efficacy(RQ4)}
\begin{table}[h!]
\centering
\caption{Ablation of Training Epoch.}
\begin{adjustbox}{width=0.45\textwidth}
\begin{tabular}{lcccc}
\toprule
& \textbf{Industry} & \textbf{MovieLens} & \textbf{KuaiRec} & \textbf{Epoch} \\
\midrule
\multicolumn{4}{@{\hspace{0.22cm}}l}{\textbf{\textit{RoBERTa}}} \\
\hdashline
Original  & 0.7784 & 0.7836 & 0.8175 & 3 \\
Double epoch & 0.7782 & 0.7837 & 0.8176 & 6 \\
Two-stage training & 0.7791 & 0.7840 & 0.8181 & 6 \\
\midrule
\multicolumn{4}{@{\hspace{0.22cm}}l}{\textbf{\textit{Qwen}}} \\
\hdashline
Original  & 0.7891 & 0.8181 & 0.8172 & 3 \\
Double epoch & 0.7887 & 0.8185 & 0.8174 & 6 \\
Two-stage training & 0.7902 & 0.8193 & 0.8189 & 6 \\

\bottomrule
\end{tabular}
\end{adjustbox}

\label{table:ablation_epoch}
\end{table}

To address RQ4 (whether the improvement from two-stage training stems from increased epochs), we compare the performance of models with original epochs, doubled epochs, and two-stage training (all with consistent total training iterations for fair comparison) on RoBERTa and Qwen architectures. The results in Table~\ref{table:ablation_epoch} reveal that the efficacy of two-stage training is independent of epoch count, directly validating the rationality of our strategy design.

For RoBERTa, the two-stage training strategy (6 epochs) outperforms both the original model (3 epochs) and the doubled-epoch model (6 epochs) across all datasets. Specifically, on the industrial dataset, its AUC (0.7791) is 0.07\% higher than the original model (0.7784) and 0.09\% higher than the doubled-epoch model (0.7782). Similarly, on KuaiRec, the two-stage strategy achieves an AUC of 0.8181, exceeding the doubled-epoch model by 0.05\%—a gain that cannot be attributed to increased training iterations, as the latter shares the same epoch count but lacks the attention-guiding mechanism.

For Qwen, two-stage training’s advantage is more pronounced. On MovieLens, its AUC (0.8193) is 0.12\% higher than the original (0.8181) and 0.08\% higher than the doubled - epoch model (0.8185). This confirms decoder architectures, lacking sink-oriented pre-training, benefit from two - stage training’s attention regulation—specifically, the first stage’s focus on [\textit{SINK}] tokens strengthens aggregating attention at behavioral boundaries, mitigating decoders’ “first-token attention bias”.

Notably, these experiments exclude the sink-specific attention enhancement mechanism, isolating the impact of the two-stage strategy. The consistent outperformance of two-stage training over both baseline and doubled-epoch models demonstrates that its effectiveness originates from actively guiding attention toward behavior-level sinks, rather than mere increases in training duration. This finding reinforces the core design principle of \textsc{CTR-Sink}: \textbf{targeted attention regulation is critical for overcoming semantic fragmentation in behavioral sequence modeling.}

\subsection{Discussion and Limitations}

Additional discussion on signal selection, attention--performance correlation, and computational limitations is provided in Appendix~\ref{appendix:discussion_limitations}.

\section{Related Work}
\subsection{LM-based CTR Prediction}

With the rapid development of LMs\cite{radford2018improving,mann2020language,zhang2022opt}, researchers start to explore the potential of PLMs for CTR prediction\cite{lin2025can,xi2024towards}. Unlike the ID-based one-hot encoding in traditional CTR prediction, the input data is transformed into textual user behaviors.

Pioneering research endeavors embraced the paradigm of In-Context Learning\cite{mann2020language,xiong2024dq}, strategically leveraging natural language prompts to formulate explicit recommendation requests that are directly fed into language models, thereby harnessing the models' inherent capacity to generate recommendation outputs\cite{zhang2021language,dai2023uncovering,gao2023chat}. 

Recently, numerous works focus on using LMs for semantic modeling of user behaviors.
Notable examples include M6-Rec\cite{cui2022m6}, which captures semantic features from user-item interactions. Tallrec\cite{bao2023tallrec} adapts large language models to recommendation tasks via tuning with recommendation data. BAHE\cite{geng_breaking_2024} uses hierarchical encoding to process long behavior sequences efficiently. CTRL\cite{li_ctrl_2023} fuses ID-based collaborative signals with LM semantic embeddings through cross-modal alignment. FLIP\cite{wang_flip_2024} performs fine-grained alignment between ID-based models and PLMs. CoLLM\cite{zhang_collm_2025} integrates collaborative information into large language models by aligning it with the LM's input space. TBHI\cite{chen_tbin_2023} and MSD\cite{zhang_balancing_2025} also contribute to encoding semantic nuances in user behaviors.

However, existing LM-based CTR models focus on feature alignment and sequence length but overlook semantic fragmentation—LMs' attention scatters across irrelevant tokens, hindering effective modeling of recommendation domain data

\subsection{Attention Sink}

The concept of attention sink is introduced by StreamingLLM\cite{xiaoefficient}, reveals that LLMs overly attend to initial tokens despite their semantic triviality. It addresses memory and computational inefficiencies in long-text scenarios, enabling better handling of long sequences.

Researchers have proposed various explanations. Some attribute it to positional biases in autoregressive training\cite{wang2024eliminating, yu2024mitigate, xiong2025dope}, where early tokens are visible to all subsequent positions , while others emphasize the role of Softmax’s requirement to normalize scores, forcing excess attention allocation to arbitrary tokens\cite{guattention, xiaoefficient,barbero2025llms}. Recent work\cite{bai_does_2024, guattention,yu2024unveiling} further observe that attention sink can also be found in some tokens with limited semantic information and non-fixed positions. These studies suggest that attention sink is related to model internal structure, parameter optimization, and data distribution.

Currently, numerous studies have begun to apply this theory to various fields. Initially, attention sink was employed in the field of NLP to enable LMs to process longer texts\cite{xiaoefficient, guattention,xiong2025parallelcomp,li2024uncertaintyrag}. Subsequent studies have also found that optimizing attention on these sinks can enhance the accuracy of LMs in various applications\cite{yu2024unveiling}. Furthermore, some works have extended its application to the multimodal domain\cite{kang2025see,acharya2024star}, effectively mitigating the hallucination problem through reallocating attention.

Though recommendation research increasingly incorporates large models, no work has applied Attention Sink theory, leading to unaddressed attention dispersion from lacking natural sinks in user behaviors, limiting modeling of behavioral correlations.

\section{Conclusion}
In this study, we tackle the semantic fragmentation challenge in LM-based CTR prediction, where discrete user behaviors (with semantically empty separators) mismatch LMs’ pre-trained attention for coherent language. We propose CTR-Sink, a framework inserting recommendation-signal-fused [\textit{SINK}] tokens between behaviors. By anchoring LM attention at behavioral boundaries, CTR-Sink not only aggregates inter-behavior dependencies but also preserves pre-trained LM capabilities. Extensive experiments on industrial and public datasets (MovieLens, KuaiRec) across RoBERTa and Qwen architectures demonstrate consistent AUC improvements of 0.2–0.5\% over baseline LM-CTR methods. In the future, we will extend CTR-Sink to multi-modal behaviors and optimize its efficiency for ultra-long sequences, advancing LM-adaptive recommendation systems.

\begin{acks}
We sincerely thank Haokun Lin from NLPR, Institute of Automation, Chinese Academy of Sciences, Zhiyuan Wu from Tsinghua University‌, Haifeng Lu from The University of Hong Kong, Zhen Zhang from Lanzhou University, Xiaolong Du from University of Science and Technology of China, Xiaoyan Yuan, Minghao Chen and Luyi Yang for inspiring suggestions.

This work was supported by Ant Group Research Intern Program.

This work was supported in part by the National Natural Science Foundation of China (Grant No. U23B2054), in part by Guangdong Province Science and Technology Foundation (No. 2026A1515011806, No. 2024TQ08X559), in part by Innovation Team Project of Guangdong Province (No. 2024KCXTD017) , in part by  Shenzhen Science and Technology Foundation (No. JCYJ20240813145816022).
\end{acks}

\clearpage

\bibliographystyle{ACM-Reference-Format}
\bibliography{sample-base}

\appendix
\section{Attention visualization of  preliminary sink
validation}
\label{appendix:cls_visualization}

\begin{figure}[htbp]
    \includegraphics[width=0.23\textwidth]{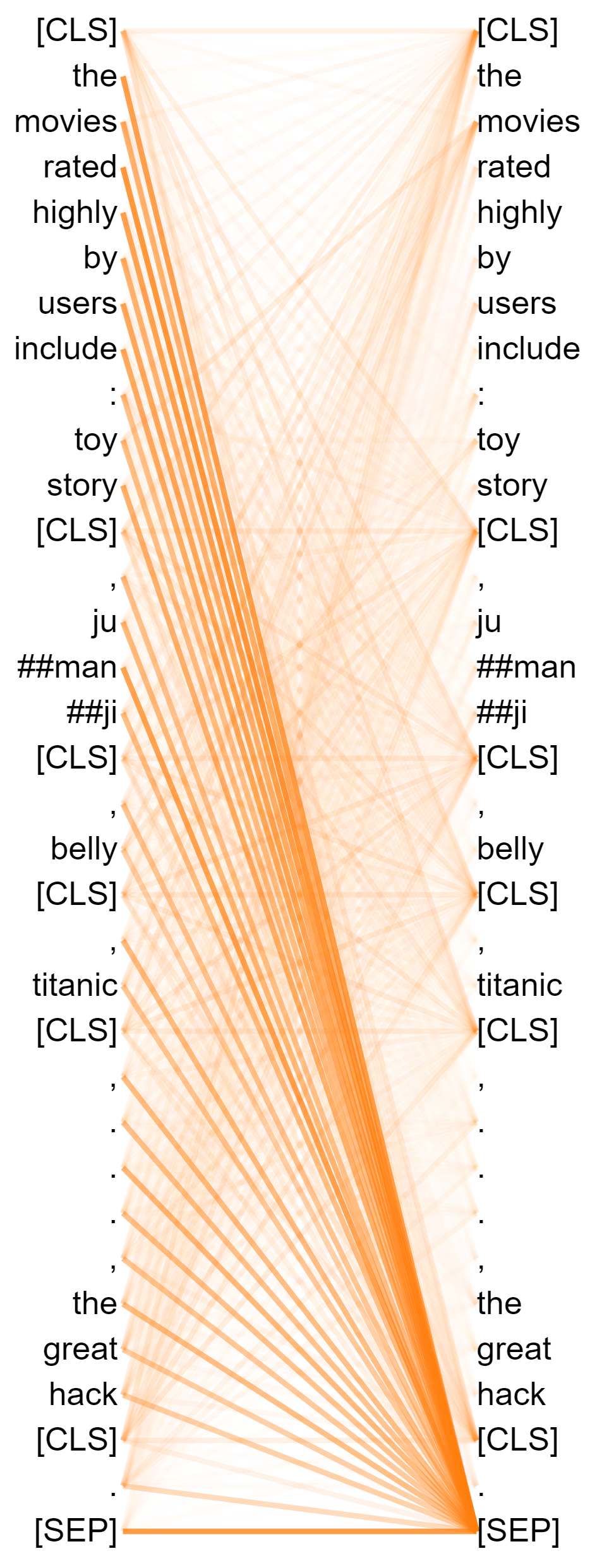}
    \caption{Attention Visualization of [CLS] Tokens as Potential Attention Sinks in Textual User Behaviors.}
    \label{cls attention}
\end{figure}

We present the attention visualization results after inserting [CLS] in Figure~\ref{cls attention}. It can be observed from the figure that each [CLS] position aggregates more attention compared to other tokens. This indicates that [CLS] can serve as an attention sink to guide the model's attention. Additionally, combined with the results in Table~\ref{table:roberta}, we find that such attention aggregation is beneficial to the model's performance.

\section{Statistics of all datasets}
\label{appendix:dataset}
In this section, we present the specific details of the datasets in Table~\ref{table:dataset}.
\begin{table}[htbp]
\centering
\caption{Statistics of all datasets.}
\begin{tabular}{llll}
\toprule
& \textbf{Industry} & \textbf{MovieLens} & \textbf{KuaiRec} \\
\midrule
Exposures  & 46,610,124 & 10,662,652 & 4,282,953 \\
Language  & Chinese & English & Chinese \\
Open source  & No & Yes & Yes \\

\bottomrule
\end{tabular}

\label{table:dataset}
\end{table}

\section{Sensitivity analysis of embedding dimension in [SINK] token generation}
\label{appendix:dim}
\begin{table}[h!]
\centering
\caption{Ablation experiments on the dimension of the embedding layer when generating [\textit{SINK}] tokens.}
\begin{tabular}{l c c c c  c} 
\toprule
Dimension   & 32 & 64 & 128 & 256  & w/o [\textit{SINK}]\\ \midrule

AUC  & 0.7830 & 0.7824 & 0.7836 & 0.7834  & 0.7808 \\
\bottomrule

\end{tabular}

\label{table:ablation_dim}
\end{table}
In this section, we analyze the sensitivity of the dimension of the embedding layer when generating [\textit{SINK}] tokens for RoBERTa, with the specific results shown in Table~\ref{table:ablation_dim}. It can be observed that the dimension of the embedding layer exerts a certain impact on the results; however, overall, the addition of [\textit{SINK}] tokens leads to a significant improvement in model performance, which strongly validates the robustness of our method.

\section{Supplementary results with larger Qwen backbones}
\label{appendix:large_qwen}

To further verify robustness under stronger decoder backbones, we conduct supplementary experiments on MovieLens with Qwen2-7B and Qwen3-8B. As shown in Table~\ref{table:large_lm_movielens}, \textsc{CTR-Sink} continues to improve both larger models, suggesting that stronger long-sequence representations do not substitute for behavior-level attention regulation.

\begin{table}[htbp]
\centering
\caption{Supplementary results with larger Qwen backbones on MovieLens.}
\begin{tabular}{llc}
\toprule
\textbf{Backbone} & \textbf{Method} & \textbf{AUC} \\
\midrule
Qwen2-7B & LM-CTR w/o \textsc{CTR-Sink} & 0.8235 \\
Qwen2-7B & LM-CTR w \textsc{CTR-Sink} & 0.8274 \\
Qwen3-8B & LM-CTR w/o \textsc{CTR-Sink} & 0.8266 \\
Qwen3-8B & LM-CTR w \textsc{CTR-Sink} & 0.8301 \\
\bottomrule
\end{tabular}
\label{table:large_lm_movielens}
\end{table}

\section{Supplementary analysis of external signals}
\label{appendix:external_signal_variants}
We further compare temporal distance with behavioral frequency and a multi-signal variant that combines temporal distance and semantic similarity. Since behavioral frequency is unavailable in some datasets, the frequency comparison is conducted only on the industrial dataset. As shown in Table~\ref{table:ablation_signal_variants}, frequency can also guide [\textit{SINK}] tokens effectively, while combining multiple signals does not consistently outperform temporal distance alone.
\begin{table}[htbp]
\centering
\caption{Supplementary comparison of external signals.}
\begin{tabular}{lccc}
\toprule
& \textbf{Industry} & \textbf{MovieLens} & \textbf{KuaiRec} \\
\midrule
Frequency & 0.7832 & -- & -- \\
Temporal & 0.7810 & -- & -- \\
\midrule
Multi-signal & 0.7804 & 0.7828 & 0.8174 \\
Temporal & 0.7810 & 0.7844 & 0.8192 \\
\bottomrule
\end{tabular}
\label{table:ablation_signal_variants}
\end{table}

\section{Supplementary analysis of longer behavior sequences}
\label{appendix:extended_length}
Due to computational resource constraints, we are unable to train Qwen-based models on sequences with more than 1,000 behaviors in the current study. We therefore extend the sequence length to 100 on the industrial dataset to further examine the trend under longer behavior contexts. As shown in Table~\ref{table:extended_length_industry}, \textsc{CTR-Sink} maintains consistent gains when the sequence length increases from 50 to 100, whereas the original model shows only marginal improvement.
\begin{table}[htbp]
\centering
\caption{Supplementary performance with longer behavior sequences on the industrial dataset using Qwen.}
\begin{tabular}{lcc}
\toprule
\textbf{Method} & \textbf{50} & \textbf{100} \\
\midrule
Original & 0.7885 & 0.7892 \\
w \textsc{CTR-Sink} & 0.7919 & 0.7937 \\
\bottomrule
\end{tabular}
\label{table:extended_length_industry}
\end{table}

\section{Additional discussion and limitations}
\label{appendix:discussion_limitations}

Temporal distance is used as the default signal mainly because it is consistently available and comparable across datasets. Frequency is also effective when available, while multi-signal fusion may introduce redundancy rather than additional attention guidance. While our visualizations and ablations suggest a link between layer-wise attention allocation and AUC gains, finer-grained correlation analysis remains future work. Due to computational resource constraints, thousand-level behavior sequences are not evaluated with Qwen-based models. As a preliminary stress test, Section~\ref{appendix:extended_length} reports 100-length industrial experiments, where \textsc{CTR-Sink} still brings consistent gains.

\end{document}